\documentclass{article} % For LaTeX2e
\usepackage{colm2024_conference}

\usepackage{microtype}
\usepackage{hyperref}
\usepackage{url}
\usepackage{wrapfig}
\usepackage{latexsym}
\usepackage[bottom]{footmisc}
\usepackage{marvosym}

\usepackage{booktabs}
\usepackage{epsfig}

\title{An Empirical Study of Catastrophic Forgetting in Large\\ Language Models During Continual Fine-tuning}

\author{
    Yun Luo \textsuperscript{\rm1*},
    Zhen Yang \textsuperscript{\rm2\thanks{Equal Contribution}},
    Fandong Meng \textsuperscript{\rm2},
    Yafu Li \textsuperscript{\rm1},
    Jie Zhou  \textsuperscript{\rm2},
    Yue Zhang \textsuperscript{\rm1,3 \textrm{\small\Letter}}
    \\
    \textsuperscript{1} School of Engineering, Westlake University, Hangzhou, 310024, P.R. China. \\
    \textsuperscript{2} Pattern Recognition Center, WeChat AI, Tencent Inc, Beijing, China.  \\
    \textsuperscript{3} Institute of Advanced Technology, Westlake Institute for Advanced Study, \\ Hangzhou, 310024, P.R. China.  \\
    \texttt{\{luoyun, liyafu, zhangyue\}@westlake.edu.cn}\\
    \texttt{\{zieenyang, fandongmeng, withtomzhou\}@tentent.com}
}

\colmfinalcopy % Uncomment for camera-ready version, but NOT for submission.
\begin{document}

\maketitle

\begin{abstract}
Catastrophic forgetting (CF) is a phenomenon that occurs in machine learning when a model forgets previously learned information while acquiring new knowledge to achieve satisfactory performance in downstream tasks. As large language models (LLMs) have demonstrated remarkable performance, it is intriguing to investigate whether CF exists during the continual instruction tuning of LLMs. This study empirically evaluates the forgetting phenomenon in LLMs' knowledge during continual instruction tuning from the perspectives of domain knowledge, reasoning, and reading comprehension. The experiments reveal that catastrophic forgetting is generally observed in LLMs ranging from 1b to 7b parameters. Surprisingly, as the model scale increases, the severity of forgetting intensifies in such a model sale range which may result from the much significant initial performance in the larger LLM. When comparing the BLOOMZ decoder-only model with the encoder-decoder model mT0, BLOOMZ exhibits less forgetting and retains more knowledge. Interestingly, we also observe that LLMs can mitigate language biases, such as gender bias, during continual fine-tuning. Furthermore, our findings indicate that 
% ALPACA maintains more knowledge and capacity compared to LLAMA during continual fine-tuning, suggesting that
general instruction tuning can help alleviate the forgetting phenomenon in LLMs during subsequent fine-tuning.
\end{abstract}

\section{Introduction} \label{sec_intro}

Large language models (LLMs) have demonstrated impressive general capabilities in tackling a wide range of tasks \citep{bubeck2023sparks}. However, when it comes to real-world applications, users often find that certain specific abilities need enhancement. In such cases, relevant task-specific data are adopted to fine-tune the model in instructions to improve its performance in particular tasks \citep{touvron2023llama,scialom2022fine}. However, the widely used LLMs such as LLAMA-Chat \citep{touvron2023llama}, ChatGPT\footnote{\href{https://openai.com/}{https://openai.com/}}, and Claude-2 \footnote{\href{https://www.anthropic.com/news/claude-2}{https://www.anthropic.com/news/claude-2}} have been fine-tuned, and continual instruction tuning in specific tasks may affect the general knowledge stored in LLMs since the original training data are inaccessible.
The study of continual instruction tuning is crucial for the practical application of LLMs.
For example, in Table \ref{examples}, an LLM is first fine-tuned for the task of title generation and then learns to generate an explanation. This scenario is related to the continual learning paradigm in machine learning, where a model is trained on a sequence of tasks \citep{lopez2017gradient,WUTONG,ke2022continual}. A significant challenge in this paradigm is catastrophic forgetting (CF), in which a model forgets previously learned knowledge due to parameter updates during the learning of new tasks, leading to decreased performance on earlier tasks \citep{li2017learning,lopez2017gradient}.

% This phenomenon raises questions about the balance between preserving the model's general understanding and adapting it to specific tasks through continual fine-tuning. 

% ine-tuning these models with instructions can further improve their alignment with human preferences \citep{ouyang2022training,stiennon2020learning}.

% Large language models are typically first pre-trained on a vast amount of text data to acquire general language understanding capabilities. Then, task-specific datasets are adopted for continual fine-tuning to adapt to particular applications or domains \citep{touvron2023llama,scialom2022fine}.
% Some studies contend that the primary function of fine-tuning is to shape the model's output format and style to better suit the target task by instruction \citep{ouyang2022training,stiennon2020learning}, rather than substantially altering its underlying knowledge. It is worth noting that as the size of the fine-tuning dataset increases, there is a potential risk of the newly acquired knowledge interfering with or overwriting the model's pre-existing knowledge. 
% This phenomenon raises questions about the balance between preserving the model's general understanding and adapting it to specific tasks through fine-tuning.
% and fine-tuning these models with instructions can further improve their alignment with human preferences \citep{ouyang2022training,stiennon2020learning}.

\cite{scialom2022fine} investigate the CF issues of T0 models during continual instruction tuning. However, their analysis primarily focuses on the forgetting phenomenon observed in downstream tasks, such as summarization and style transfer. The evolution of general knowledge stored in the original pre-trained LLMs during instruction tuning remains unexplored. \cite{luo2023investigating} conduct an analysis using probing methods on pre-trained language models to examine the problem of generality destruction and general knowledge forgetting during continual fine-tuning. Nevertheless, their study is restricted to encoder-only models and classification tasks.
In this work, we draw attention to the following fundamental questions regarding forgetting in generative LLMs:

  \begin{enumerate}
  \vspace{1mm}
    \item \textit{Are the general knowledge stored in LLMs forgotten during continual instruction tuning?}
     % \item \textit{What types of general knowledge are saliently forgotten?}
      \vspace{2mm}
      \item \textit{What are the effects of model scales, model architectures, and general instruction tuning in the forgetting problem?}
      \vspace{2mm}
      \item \textit{How to mitigate such forgetting phenomenon?}
      \vspace{1mm}
     % \item \textit{Are the forgetting problem the same in different model architectures such as encoder-decoder models and decoder-only models?}
\end{enumerate}

To address these questions, we conduct an empirical study on various LLLMs, such BLOOMZ, mT0 \citep{muennighoff2022crosslingual}, LLAMA \citep{touvron2023llama}, and ALPACA \citep{alpaca}  to analyze the catastrophic forgetting (CF) problem during continual instruction tuning. We continually train the original LLMs with five instruction tasks and evaluate the retention of the general knowledge within the model from three perspectives: domain knowledge, reasoning, and reading comprehension. Furthermore, we investigate the evolution of bias in LLMs throughout the tuning process. To gain insights into the effect of model architecture, we compare the performance of BLOOMZ with that of mT0 \citep{muennighoff2022crosslingual} (an encoder-decoder model), which is fine-tuned using similar datasets. We also investigate the impact of general instruction tuning on the CF problem by comparing the performance of the initial model with the instruction-tuned version such as (BLOOM, BLOOMZ) and (LLAMA \citep{touvron2023llama}, ALPACA \citep{alpaca}).
\begin{table*}[t]\small
    \caption{Fine-tuning examples in an instruction form.}
    \centering
    \begin{tabular}{p{0.9\textwidth}}
        \hline
\textbf{Instruction}: Make a title for this article with `computer': japan's NEC corp. and UNK computer corp. of the United States said Wednesday they had agreed to join forces ...  \\
% \hline
\textbf{Output}: NEC UNK in computer sales tie-up.  \\
\hline
\textbf{Instruction}: Explain why the two following sentences are unrelated: Sentence 1: `This church choir sings to the masses as they sing joyous songs from the book at a church'; Sentence 2: `The church has cracks in the ceiling.'  \\
% \hline
\textbf{Output}: There is no indication that there are cracks in the ceiling of the church. \\ \hline
\textbf{Instruction}: The del operator shows up in the Laplacian, which is in the kinetic energy term, when the Schrodinger equation is written in the position basis. 2207\^(2) is a differential operator, equal to (d/dx)\^(2) + (d/dy)\^(2) + (d/dz)\^(2), in Cartesian coordinates (the derivatives are partial derivatives).h\^(2)2207\^(2)/2m is the kinetic energy operator in the position basis.=== Given the above text, write the possible curious question it answers:  \\
% \hline
\textbf{Output}: What is the del operator in the Schrodinger equation? \\\hline
\textbf{Instruction}: Reformulate this text with simpler words: `His father Robert Alda -LRB- born Alphonso Giuseppe Giovanni Roberto D 'Abruzzo -RRB- was an actor and singer, and his mother Joan Browne was a homemaker and former beauty-pageant winner. ' \\ 

\textbf{Output}: His father, Robert Alda, was an actor, and his mother, Joan, was once crowned Miss New York.  \\
\hline
    \end{tabular}
    % \vspace{-2mm}
    \label{examples}
    % \vspace{-5mm}
\end{table*}

Our findings reveal that the forgetting problem is generally present in LLMs. Surprisingly, as the model scale increases from 1b to 7b parameters, the severity of forgetting intensifies. One potential explanation for this phenomenon is that larger language models exhibit stronger initial performance and, consequently, experience more pronounced performance degradation because of the fitting on the new task during continual instruction tuning. Additionally, we observe that the bias in LLMs is mitigated throughout the continual instruction tuning process. When comparing BLOOMZ with mT0 at comparable model scale, we find that BLOOMZ experiences a relatively milder forgetting problem, suggesting that the decoder-only architecture may be better at retaining information during continual instruction tuning. Lastly, empirical results on LLAMA and its instruction-tuned version (i.e., ALPACA) indicate that diverse instruction tuning can help alleviate the CF phenomenon for LLMs in further continual fine-tuning.

The contribution of our paper can be summarized as follows:
\begin{enumerate}
    \item We take an initial step to analyze the catastrophic forgetting (CF) problem during continual instruction tuning by an empirically study, where a specific evaluation setting is designed from the perspective of general knowledge such as domain knowledge, reasoning, reading comprehension and the bias problem.
    \item We provide an initial research evidence that the CF problem generally exists in the continual instruction tuning process for different models such as BLOOMZ, mT0, LLAMA and ALPACA. We also show that the model architecture, and model scale  have different effects on the CF problem.
    \item Experimental results further show that the general instruction data can help mitigate the CF problem to some extent by experiments.
\end{enumerate}

\section{Related Work}
\subsection{Instruction Tuning}
Instruction tuning has proven to be effective in aligning responses from pre-trained language models with human intents or preferences \citep{ouyang2022training,stiennon2020learning,min2021metaicl}. This technique refines a model's ability to predict a specific response to a given prompt, which may optionally include an instruction that outlines a task for the model. Examples of such models include T0 \citep{sanh2021multitask}, mT0 \citep{muennighoff2022crosslingual}, and BLOOMZ \citep{muennighoff2022crosslingual}. It has been demonstrated that instruction tuning can enhance the ability of language models to generalize to unseen tasks without prior exposure \citep{wei2021finetuned,sanh2021multitask}. In this work, we focus on fine-tuning LLMs in a continual manner and analyze the catastrophic forgetting (CF) phenomenon during training. Specifically, instructions for a particular type of task (such as generating headlines) are used to tune the LLMs in each training phase, and the model does not have access to previously learned tasks.

\subsection{Evaluation of CF in Continual Learning}
Various training strategies have been proposed to address the problem of catastrophic forgetting (CF) in continual learning \citep{riemer2019learning,buzzega2020dark,ke2022a,chen2022adaprompt,luo2023mitigating}. Previous studies have primarily measured CF by evaluating the performance decrease in previously learned tasks during continual learning or the average performance of learned tasks at the end of training. However, \cite{davari2022probing} discovered that even when the model performance on previously learned tasks is preserved, the representations still suffer from significant drift due to parameter updates. As a result, they propose using an optimal linear classifier of learned tasks to measure performance, with changes considered as a surrogate to quantify CF. Similarly,  \cite{WUTONG} employs layer-wise and task-wise probing to analyze CF in each layer for previously learned tasks.  \cite{luo2023investigating} propose using a series of probing tasks to evaluate the knowledge stored in LLMs and analyze the generality of the models. However, their study is limited to classification tasks and encoder-only model architectures. To the best of our knowledge, we are the first to evaluate the forgetting of general knowledge in generative large language models during continual instruction tuning.

\section{Method}
\begin{figure*}
    \centering
    \includegraphics[width=\hsize]{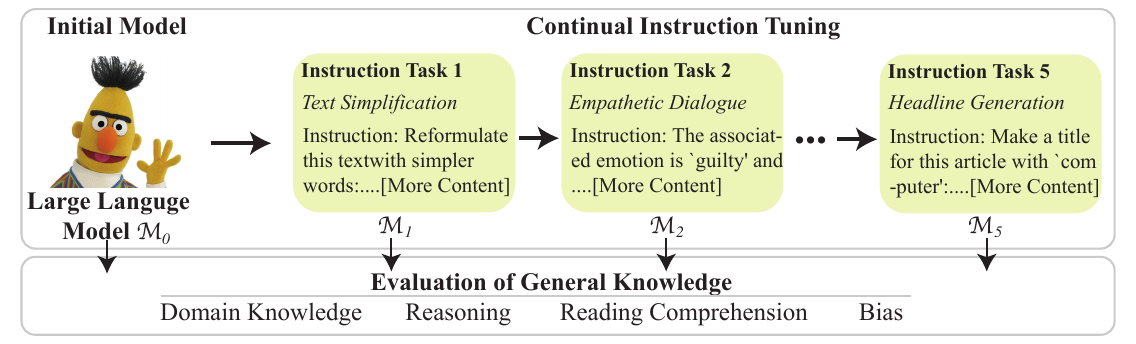}
    % \vspace{-2mm}
    \caption{The framework in our empirical study for the continual instruction tuning. The initial model $M_0$ is continually trained with different instruction tasks and evaluated from the perspective of general knowledge tasks including domain knowledge, reasoning, reading comprehension, and the problem of bias.}
    \label{framework}
        % \vspace{-2mm}

\end{figure*}

Formally, in the continual instruction tuning of LLMs, a model sequentially learns several generation tasks denoted as $\mathcal{T} = \{T^m\}, m = 1,2,...,N$ ($N$ is the length of the task sequence).  During the training of each task $T^m \in \mathcal{T}$, only the corresponding data $D^m = \{(x^m_i,y^m_i)\}$ are available, where $x^m_i$ is the input text together with an instruction and $y^m_i$ is the corresponding generation labels. Given an initial LLM denoted by $M_0$, we continually train the model with the data $D^m$, obtaining the trained model $M_m$. The training and evaluation framework is shown in Figure \ref{framework}.

\begin{table*}[]\small
\caption{Details of the evaluation sets for the CF phenomenon in LLMs. DK, Rs, and RC represent domain knowledge, reasoning, and reading comprehension.}
\centering
\begin{tabular}{ll}
\hline  
Set & Elements       \\ \hline
DK      &  STEM, Social, Human, Other \\
Rs      &  BoolQ, PIQA, Winogrande, Hellaswag, MathQA, Mutual \\
RC & RACE-high, RACE-middle \\
Bias & Sexual Orientation, Physical Appearance,  Religion, Nationality, Race/Color, Gender,  \\ 
&Socioeconomic, Disability, Age\\

\hline
\end{tabular}
% \vspace{-2mm}

\label{eval_t}
    % \vspace{-3mm}

\end{table*}

\subsection{Continual Tasks}
Following Scialom et al. (2022) \citep{scialom2022fine}, we adopt instruction tasks dissimilar to the training and evaluation tasks of BLOOMZ and mT0. Specifically, we select 5 tasks from Scialom et al. (2022) as follows: 
\begin{enumerate}
    \item \textbf{Text Simplification (Simp)} \citep{jiang2020neural,alva2020asset} requires paraphrasing the text with a simple text;
    
    \item \textbf{Empathetic Dialogue Generation (Emdg)} \citep{rashkin2019towards}, requires the model to generate a reason for a  conversational context under a given emotional situation;

\item \textbf{Inquisitive Question Generation (InqQG)} \citep{fan2019eli5} requires the model to generate a question for the long-form answers;

\item \textbf{Explanation Generation (Exp)} \citep{camburu2018snli},  aims to train a model able to generate natural language explanations for a given premise, hypothesis, or label;

\item \textbf{Headline Generation with Constraint (HGen)} \citep{scialom2022fine} aims to train a model able to generate headlines under some specific constraints, such as containing the keywords X at the beginning, at the ending, or anywhere.
\end{enumerate}

During the instruction tuning, we first add a general prompt template to the beginning of the data:
\textit{`Below is an instruction that describes a task, paired with an input that provides further context. Write a response that appropriately completes the request...'} followed by a specific prompt for each task. We adopt the specific prompts designed by Scialom et al. (2022) \citep{scialom2022fine} and 100,000 data samples are used for training.  The details of the instruction are shown in Appendix \ref{instruction_detail}. For simplicity, we train the model on one instruction task order for the empirical study: Simp $\to$ Emdg $\to$ InqQG $\to$ Exp $\to$ HGen.

\subsection{Evaluation Tasks}
% In order to evaluate the general/basic knowledge stored in the LLMs, we adopt several general evaluation tasks, which can be categorized into four sets:
% 1) \textbf{Domain Knowledge}, we adopt massive multitask language understanding benchmark (MMLU) \citep{hendrycks2020measuring} to evaluate the knowledge stored in the LLMs, which can be divided into STEM, Human, Social, and Other; 2) \textbf{Reasoning},  we adopt the commonly used commonsense reasoning datasets, Hellaswag \citep{zellers2019hellaswag}, BoolQ \citep{clark2019boolq}, Winogrande \citep{sakaguchi2021winogrande}, PIQA \citep{bisk2020piqa}; math reasoning datasets mathQA \citep{amini2019mathqa}; dialog reasoning dataset Mutual \citep{cui2020mutual}; 3) \textbf{Reading Comprehension}, we evaluate the LLMs on the dataset RACE \citep{lai2017race} for both middle and high split; 4) \textbf{Bias}, we also evaluate the biases in the continually trained model on the CrowSPairs \citep{nangia2020crows}, which evaluates biases including gender, race/color, religion, etc.
To evaluate the general/basic knowledge stored in the LLMs, we adopt several general evaluation tasks (Table \ref{eval_t}), which can be categorized into four sets:

\textbf{Domain Knowledge}: We employ the Massive Multitask Language Understanding benchmark (MMLU) \citep{hendrycks2020measuring} to assess the knowledge stored in the LLMs. MMLU covers a wide range of domains, including STEM, Human, Social, and Other.

\textbf{Reasoning}: We utilize commonly used commonsense reasoning datasets, such as Hellaswag \citep{zellers2019hellaswag}, BoolQ \citep{clark2019boolq}, Winogrande \citep{sakaguchi2021winogrande}, and PIQA \citep{bisk2020piqa}. Additionally, we evaluate the models on mathQA \citep{amini2019mathqa} for math reasoning and Mutual \citep{cui2020mutual} for dialog reasoning.

\textbf{Reading Comprehension}: We assess the LLMs' performance on the RACE dataset \citep{lai2017race}, which includes both middle and high school level reading comprehension tasks.

\textbf{Bias}: To investigate the biases in the continually trained models, we employ the CrowSPairs dataset \citep{nangia2020crows}, which evaluates various biases, including gender, race/color, religion, and more.

\subsection{Evaluation Metric}
Formally, we define $\mathcal{E} = \{E_i\}, i = 1,2,3,4$, as the above different evaluation sets, where each set contains different datasets or different splits (Table \ref{eval_t}). For example, $E_1$ refers to the evaluation set of MMLU, and it contains four elements -- STEM, Human, Social, and Other. For each element $e \in E_i$, we adopt $R^e_m$ as the evaluation results, where $m$ refers to the order of continually trained tasks,  i.e. the number of fine-tuning tasks the model has been continuously trained on.

We define the forgetting metric $FG$, which is the average decrease of $R^e_m$, as a surrogate metric to evaluate the forgetting:
\begin{equation}
\centering
    FG_i = \frac{1}{|E_i|}\sum_{e\in E_i} \frac{1}{N} \sum_{m=1}^{N} \frac{R^{e}_o-R^{e}_m}{R^{e}_o} *100\%,
\end{equation}
where ${R^{e}_o}$ is the results of $e$ on initial LLMs.  We obtain the evaluation results using the open-source evaluation framework -- \textit{lm-evaluation-harness} \citep{eval-harness}.

For most multi-choice evaluation elements, we adopt the accuracy in the zero-shot setting to measure the model performance, including MathQA, Hellaswag, BoolQ, PIQA, Mutual, Winograde, and RACE. For MMLU, we adopt the 5-shot setting for evaluation.
For CrowsPairs, we follow Nangia et al. (2022) \citep{nangia2020crows} to 
measure the model preference for the stereotypical sentence based on the perplexity of the given stereotypical and anti-stereotypical sentences, where a larger value means a stronger bias in the language.

\section{Experimental Setting}
\subsection{Large Language Models}
We adopt BLOOMZ for the empirical study since BLOOMZ is diverse in the scales and can be directly compared with the encoder-decoder model mT0, which is fine-tuned on the same instruction datasets as BLOOMZ. We also consider the widely used LLAMA and ALPACA to further study the effect of general instruction tuning.

\noindent\textbf{BLOOMZ} \citep{muennighoff2022crosslingual} is a decoder-only model fine-tuned on multilingual tasks with English prompts based on the model BLOOM. The model scales range from 560M to 176b, providing a test bed for analyzing the forgetting phenomenon across different scales.  Specifically, in this study, we continually train BLOOMZ on the scale 1.1b, 1.7b, 3b, and 7.1b \footnote{\href{https://huggingface.co/bigscience/bloomz}{https://huggingface.co/bigscience/bloomz}} because of the limitation of computation resources.

\noindent\textbf{mT0} \citep{muennighoff2022crosslingual} is a encoder-decoder model based on T5. The model is fine-tuned on similar tasks as BLOOMZ. Specifically, we adopt the scale of 1.2b and 3.7b for comparison with BLOOMZ to analyze the effect of model architecture in the CF problem. \footnote{\href{https://huggingface.co/bigscience/mt0-xl}{https://huggingface.co/bigscience/mt0-xl}}

\noindent\textbf{LLAMA} \citep{touvron2023llama}  is an open-source decoder-only model based on publicly available data, and achieves competitive results compared with the existing LLMs.  \footnote{\href{https://huggingface.co/decapoda-research/llama-7b-hf}{https://huggingface.co/decapoda-research/llama-7b-hf}}

\noindent\textbf{ALPACA} \citep{alpaca} is a model fine-tuned on LLAMA-7b using 52K instruction data generated by the techniques in the Self-Instruct \citep{selfinstruct}. And ALPACA behaves similarly to text-davinci-003 on the Self-Instruct instruction-following evaluation suite. \footnote{\href{https://huggingface.co/tatsu-lab/alpaca-7b-wdiff}{https://huggingface.co/tatsu-lab/alpaca-7b-wdiff}}

\subsection{Implementation}
 We train our model on 8 GPU (Tesla A100 40G) using the Adam optimizer \citep{kingma2014adam} (the models in 1b level are trained on 4 GPU for saving resources). For all the models, the batch size is 4 on each device,  the learning rate is 2e-5, and the scheduler is set constant for BLOOMZ and mT0 following \citep{muennighoff2022crosslingual}. In LLAMA and ALPACA, we follow the hyperparameter of Taori et al. (2023) \citep{alpaca} that the scheduler is cosine and learning rate is 2e-5. \footnote{Note that these hype-parameters are set according to the initial work, but may still introduce some effects in the fine-tuned performance.}  The max sequence length of the inputs is 512. We train our model 3 epochs and the final checkpoints are used for evaluation.

\section{Experimental Results and Analysis}
In this section, we first show that the forgetting phenomenon generally exists in LLMs during continual instruction tuning in Section \ref{mainresultssec}. Then we analyze the factors that affect the forgetting extent, such as model scales, model architectures, and general instruction tuning in Section \ref{scalesec}-\ref{inssec}, respectively.

\begin{table*}[]\scriptsize
\caption{The performance of some LLMs before and after instruction tuning on the corresponding task in the continual learning. `Initial' refers to the performance of the original LLMs. `Tuned' refers to the performance after instruction tuning on this task.  R1, and BS
denote  ROUGE-1 and BERTScore, respectively.}
\centering
\vspace{2mm}
% \resizebox{\columnwidth}{!}{%
\begin{tabular}{l|cc|cc|cc|cc|cc}
\hline
           & \multicolumn{2}{c|}{Simp (SARI)} & \multicolumn{2}{c|}{Emdg (BS)} & \multicolumn{2}{c|}{InqQG (BS)} & \multicolumn{2}{c|}{Exp (BS)}& \multicolumn{2}{c}{HGen (R1)} \\
           % \hline 
           & Initial &Tuned &Initial &Tuned &Initial &Tuned&Initial &Tuned&Initial &Tuned\\ \hline
mT0-3.7b &39.01&39.92&48.29&51.70&52.66&56.25&51.62&61.91&30.50&31.77 \\
% BLOOMZ-1.7b &37.18&&47.31&&49.50&&50.43&&28.13 \\
BLOOMZ-3b &37.95&46.72&46.10&53.27&49.06&59.72&49.75&67.10&27.88&31.72 \\
BLOOMZ-7.1b &45.26&47.24& 49.68&53.30&52.30&59.69&51.47&68.71&31.50&32.93 \\
BLOOM-7.1b &42.65&47.14&44.98&52.37&44.30&59.99&49.57&68.76&30.50&32.41 \\
LLAMA-7b &43.02&46.92&44.28&49.54&43.90&47.54&50.72&54.22&32.42&33.80\\
ALPACA-7b &45.37&48.22&52.56&54.70&56.91&62.13&52.66&70.49&32.06&36.73 \\
\hline
\end{tabular}

\label{tuning}
\end{table*}

\begin{table*}[t]\small
\centering
\caption{Main results of the forgetting in LLMs during continual instruction tuning. $R^s_{o}$ and $R^s_{-1}$ refer to the evaluation results at the beginning and the end of instruction tuning.}
\begin{tabular}{l|ccc|ccc|ccc}
\hline
           & \multicolumn{3}{c|}{Domain Knowledge} & \multicolumn{3}{c|}{Reasoning} & \multicolumn{3}{c}{Reading Comprehension}  \\
           \hline
    & $R^e_o$       & $R^e_{-1}$        & $FG$      & $R^e_o$        & $R^e_{-1}$       & $FG$      & $R^e_o$      & $R^e_{-1}$    & $FG$        \\ \hline
mT0-1.2b  &26.82 &22.47&9.18&45.43&40.22&7.75&35.06&29.54&17.45 \\
mT0-3.7b     & 30.99         &   20.14       &    20.15     &  48.61        & 38.39       & 16.73         &  41.10      &   30.45    & 28.42            \\
BLOOMZ-1.1b &  27.19        & 23.84         & 9.54        &    47.37      &   41.97       &   6.73      &   36.77     & 27.28      &   18.04          \\
BLOOMZ-1.7b &  28.72        &  24.52        &   10.72      &   48.30       &   44.96       &   6.48      &   42.65     &  30.09     & 24.29           \\
BLOOMZ-3b  &  30.04        &  24.29        &    14.63     &  56.17        &   47.03       &   11.09       &  48.29      &  31.38    & 27.56           \\
BLOOMZ-7.1b  &   33.08       &  25.61        &   18.37      &  59.15        &   49.24       &   13.62       & 48.79         &  33.05     &   26.75      \\
% LLAMA-7b &37.27&24.05&34.57&58.73&40.38&31.33&41.36&27.62&31.72 &56.29&50.35&10.11\\
% ALPACA-7b  &  39.29 & 29.88&18.14& 60.11&53.68&7.56&44.47&37.61&10.31&62.44&66.49&-3.63\\
\hline
\end{tabular}

\label{main}
    % \vspace{-3mm}

\end{table*}

\subsection{Main Results}
\label{mainresultssec}

\begin{wrapfigure}[15]{r}{0.5\textwidth}
    \centering
    \includegraphics[width=0.47\textwidth]{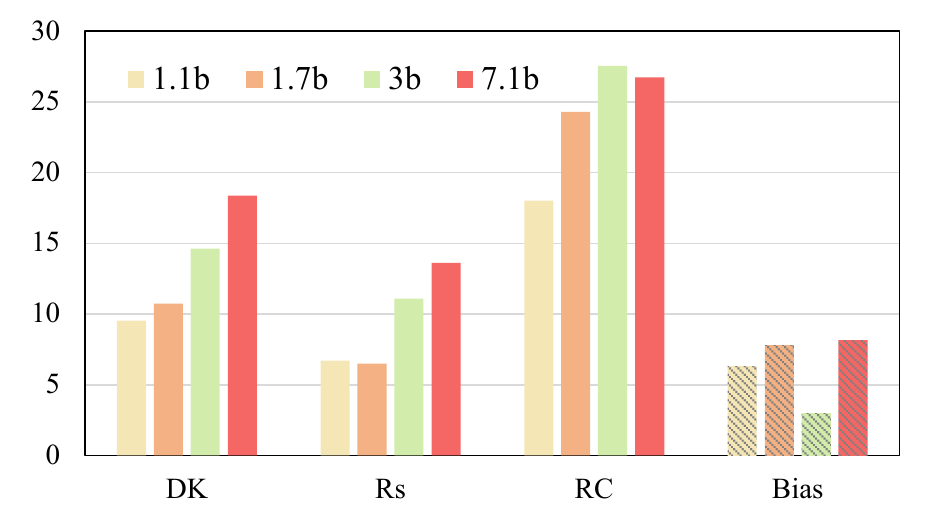}
    % \vspace{-2mm}
    \caption{The FG values of BLOOMZ in different model scales after continually training.}
    \label{tuned_bloomz}
\end{wrapfigure}

Firstly, we show the results of the instruction tuning during continual instruction tuning (Table \ref{tuning}). We mainly present the initial performance and the performance after tuning on the corresponding task to demonstrate the effectiveness of instruction tuning.  We use the metrics to measure the model performance on each task following Scialom et al. (2022) \citep{scialom2022fine}. In particular, we use SARI for Simp, BLUEScore for Emdg, InqQA, Exp, and Rouge-1 for HGen.  For example, the initial BLOOMZ-7.1b achieves 51.47\% BLUEScore in the \textit{Exp} task, but after continual tuning the model on Simp, Emdg, InqQG, and Exp, the model achieves 68.71\% on the Exp task. These improvements demonstrate that the model can benefit from the instruction tuning process and achieve significantly better performance on the instruction tasks.

% Firstly, Figure \ref{tuned_bloomz} shows the performance of the continually trained BLOOMZ-7.1b model on the instruction tasks compared to the initial model's performance. For example, after being trained on the Simp training data, the BLOOMZ model achieves a SARI score of 47.24 on the Simp test set, which is 1.98 higher than the initial model's score. Subsequently, when we continually train the model on the Emdg dataset, the obtained model achieves a BLEU score of 53.3, surpassing the initial model's performance by 3.62. 
% Additional results for other LLMs are presented in Appendix \ref{InstuctionResults}.

\begin{wraptable}{r}{0.5\textwidth} \small
\caption{FG values for Bias in LLMs during continual instruction tuning. $R^s_0$ and $R^s_{-1}$ refer to the evaluation results at the beginning and the end of instruction tuning.}
\centering
\begin{tabular}{lccc}
\hline
& $R^s_0$ & $R^s_{-1}$ & $FG$ \\
\hline
mT0-1.2b & 56.31 & 53.46 & 5.62 \\
mT0-3.7b & 57.16 & 50.59 & 13.10 \\
BLOOMZ-1.1b & 61.07 & 58.65 & 6.27 \\
BLOOMZ-1.7b & 65.18 & 56.48 & 7.78 \\
BLOOMZ-3b & 63.90 & 62.14 & 2.97 \\
BLOOMZ-7.1b & 65.82 & 60.61 & 7.15 \\
\hline
\end{tabular}
\label{bias}
\end{wraptable}

\begin{table*}[t]\small
\caption{Results of CF in the models w/o general instruction tuning, including  the pairs (BLOOM, BLOOMZ), and (LLAMA, ALPACA).}
\centering
\begin{tabular}{l|ccc|ccc|ccc}
\hline
           & \multicolumn{3}{c|}{Domain Knowledge} & \multicolumn{3}{c|}{Reasoning} & \multicolumn{3}{c}{Reading Comprehension}\\
           \hline
    & $R^e_o$       & $R^e_{-1}$        & $FG$      & $R^e_o$        & $R^e_{-1}$       & $FG$      & $R^e_o$      & $R^e_{-1}$    & $FG$      \\ \hline
BLOOM-7.1b &29.42&24.83&13.54&52.79&47.76&6.67&38.25&31.55&12.00\\
BLOOMZ-7.1b  &   33.08       &  25.61        &   18.37      &  59.15        &   49.24       &   13.62       & 48.79         &  33.05     &   26.75        \\
LLAMA-7b &37.27&24.05&34.57&58.73&40.38&31.33&41.36&27.62&31.72\\
ALPACA-7b  &  39.29 & 29.88&18.14& 60.11&53.68&7.56&44.47&37.61&10.31\\
\hline
\end{tabular}
% \vspace{-2mm}
\label{tuned}
    % \vspace{-4mm}

\end{table*} 

% \begin{table}[]\small
% \vspace{3mm}
%     \caption{FG values for Bias in LLMs during continual instruction tuning. $R^s_{o}$ and $R^s_{-1}$ refer to the evaluation results at the beginning and the end of instruction tuning.}
%     \centering
% \begin{tabular}{lccc}
% \hline
%     &$R^e_o$      & $R^e_{-1}$     & $FG$        \\
%     \hline
% mT0-1.2b & 56.31&53.46&5.62  \\
% mT0-3.7b & 57.16&50.59&13.10 \\
% BLOOMZ-1.1b &61.07&58.65&6.27\\
% BLOOMZ-1.7b &65.18&56.48&7.78\\
% BLOOMZ-3b &63.90&62.14&2.97\\
% BLOOMZ-7.1b &65.82&60.61&7.15\\

% \hline
%     \end{tabular}

%     \label{bias}
% \end{table}？

% \begin{figure}
%     \centering
%     \includegraphics[width=0.5\textwidth]{it_1.pdf}
%     % \vspace{-8mm}
%     \caption{The FG values of BLOOMZ in different model scales after continually training. }
%     \label{DK}
% \end{figure}

\begin{figure*}
    \centering
    \includegraphics[width=\hsize]{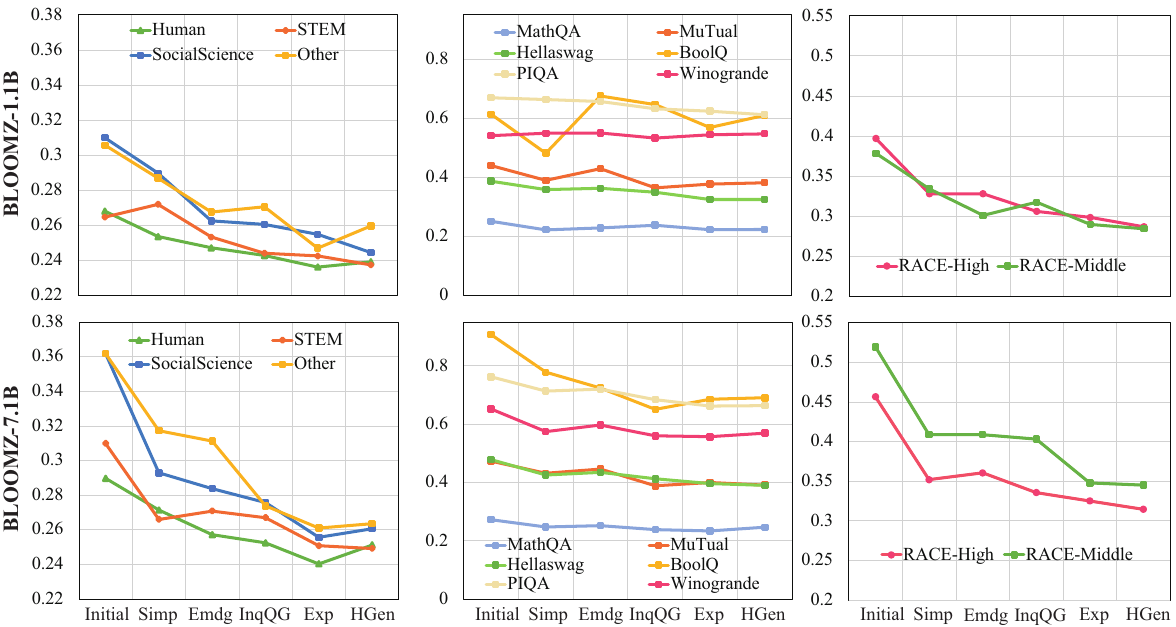}
    % \vspace{-3mm}
    \caption{The evaluation of detailed performance in BLOOMZ-1.1b and BLOOMZ-7.1b during continual instruction tuning. The first row refers to the model BLOOMZ-1.1b and the second refers to BLOOMZ-7.1b. The first to third columns are the FG values of domain knowledge (MMLU), reasoning and reading comprehension, respectively.}
    \label{details}
        % \vspace{-5mm}

\end{figure*}

Next, Figure \ref{details} displays the FG values of BLOOMZ-1.1b and BLOOMZ-7.1B. As we can observe, the performance gradually decreases as we continually tune the model with instruction tasks. For instance, the performance of BLOOMZ-7.1b on MMLU-SocialScience in Figure \ref{details} drops from 36.18\% to 26.06\% after continual training. The declining performance in LLMs indicates the presence of the catastrophic forgetting (CF) problem during the continual instruction tuning process. Moreover, as more instruction tasks are introduced, the general knowledge suffers more significant forgetting.
We can also notice that the performance of the BLOOMZ-7.1b model drops more drastically in these evaluation tasks, as we adopt the same y-axis scale for both BLOOMZ-1.1b and BLOOMZ-7.1b in Figure \ref{details}. For example, the performance on MMLU-Other drops from 36.18\% to 26.35\% in BLOOMZ-7.1b, while it drops from 30.58\% to 25.97\% in BLOOMZ-1.1b after continual training. Although the initial performance of BLOOMZ-7.1b is more substantial compared to that of BLOOMZ-1.1b, the performance of both models ends at relatively similar values, which would result in a more significant FG value for BLOOMZ-7.1B.
% \footnote{Detailed results of some other LLMs during continual instruction tuning are presented in the Appendix \ref{others}.}

%   \begin{figure*}
%     \centering
%     \includegraphics[width=0.76\hsize]{architecture.pdf}
%     % \vspace{-3mm}
%     \caption{The  comparison of FG values between BLOOMZ and mT0 in the comparable model scale. DK, Rs and RC refer to the abbreviations of domain knowledge, reasoning and reading comprehension.}
%     \label{arcdetailed}
%     % \vspace{-1mm}
% \end{figure*}

  \begin{figure*}
    \centering
    \includegraphics[width=\hsize]{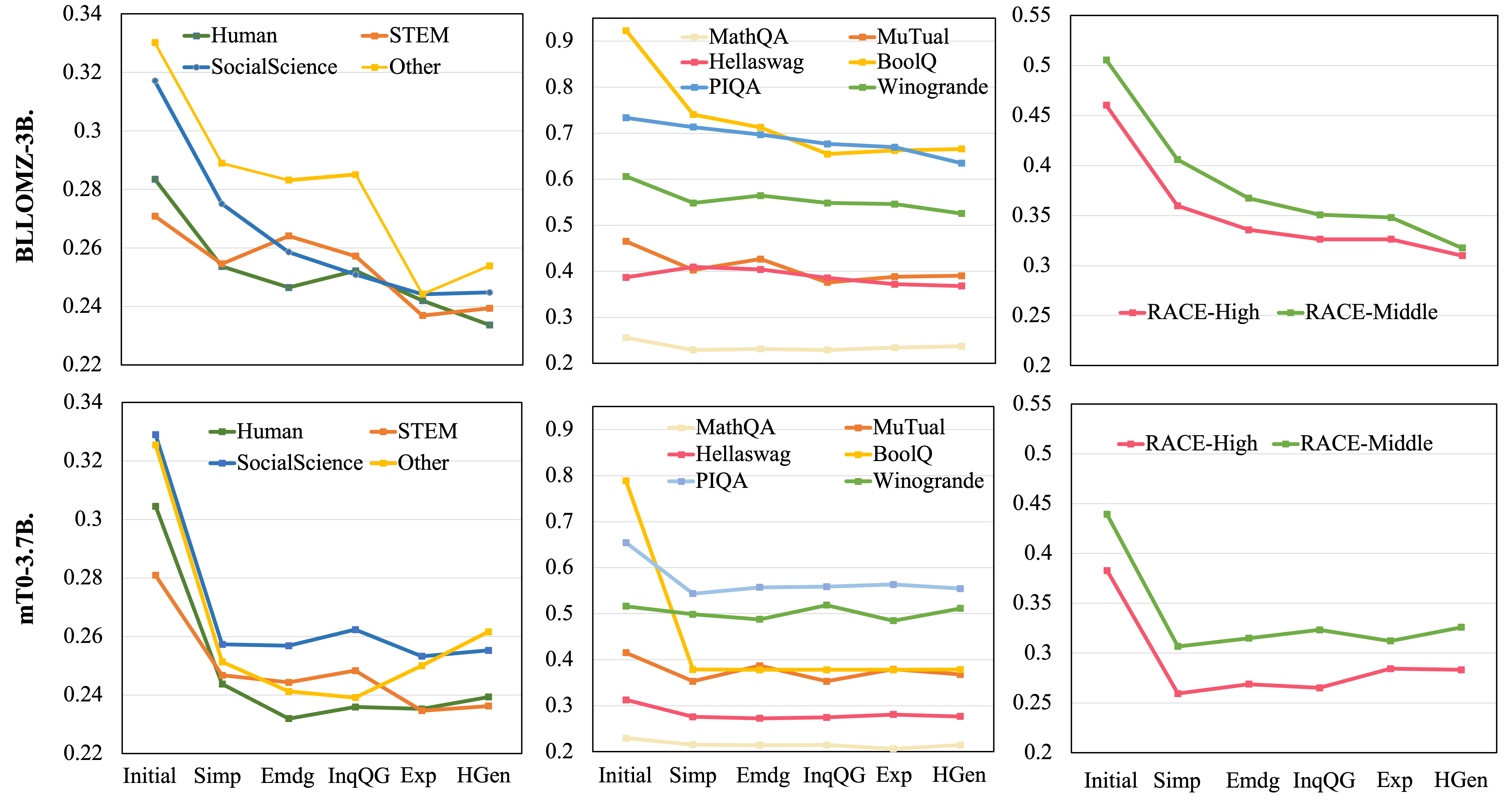}
    % \vspace{-3mm}
    \caption{The detailed results of knowledge evolution between BLOOMZ and mT0 in the comparable model scale. The first row refers to the model BLOOMZ-3b and the second refers to mT0-3,7b. The first to third columns are the FG values of domain knowledge (MMLU), reasoning and reading comprehension, respectively.}
    \label{arc}
    % \vspace{-1mm}
\end{figure*}

\begin{figure*}
    \centering
    \includegraphics[width=\hsize]{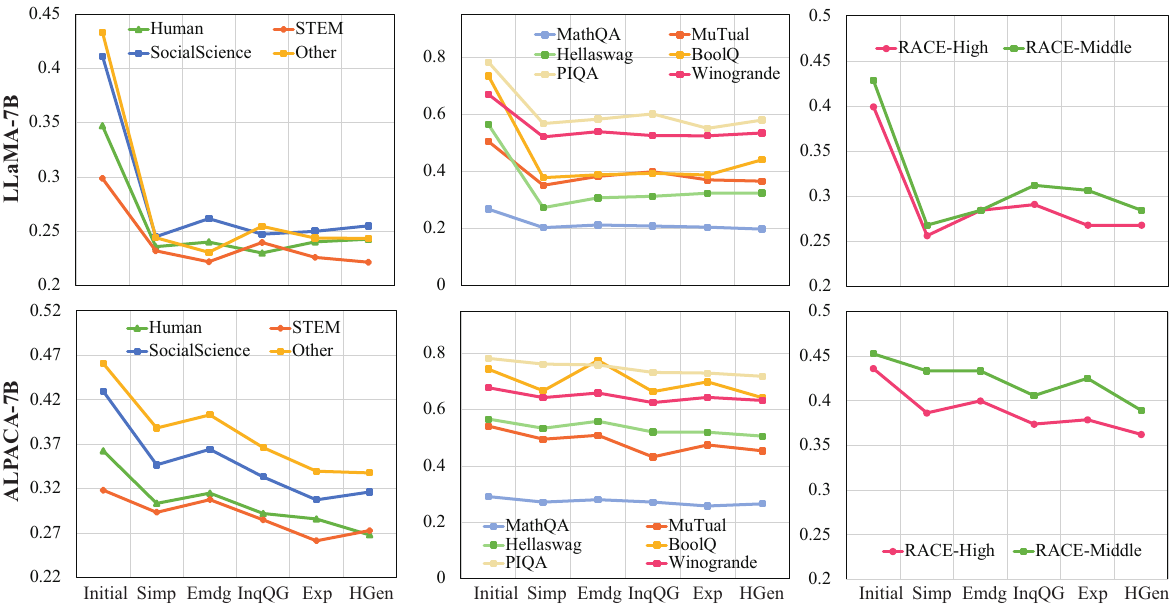}
    % \vspace{-4mm}
    \caption{The evaluation of detailed performance in LLAMA-7b and ALPACA-7b during continual instruction tuning. The first row refers to the model LLAMA-7b and the second refers to ALPACA-7b. The first to third columns are the FG values of domain knowledge (MMLU), reasoning and reading comprehension, respectively.}
    \label{llama_details}
        % \vspace{-3mm}
    
\end{figure*}

% The main results of forgetting are reported in Table \ref{main}. We observe that FG values of domain knowledge, reasoning, and reading comprehension are all above zero, indicating that general knowledge is forgotten during continual instruction tuning. The performance of reading comprehension suffers the most drastic forgetting, followed by domain knowledge. For example, FGs of BLOOMZ-7.1b are 26.75\%,  18.37\%, and 13.62\%  in reading comprehension, domain knowledge, and reasoning, respectively. 
% However, we observe that FGs in Bias (Table \ref{bias}) are mostly above zero in the experiments, which indicates that the model biases, such as race color, gender, and so on, are mitigated during continual instruction tuning. For example, in the sentences describing physical appearance, BLOOMZ-7.1b prefers the stereotype ones with a probability of 75.0\% at the beginning but achieves 63.88\% after continual instruction tuning. 
The main results of forgetting are reported in Table \ref{main}. We observe that the FG values for domain knowledge, reasoning, and reading comprehension are all above zero, indicating that general knowledge is forgotten during continual instruction tuning. Reading comprehension performance suffers the most drastic forgetting, followed by domain knowledge. For example, the FG values of BLOOMZ-7.1b are 26.75\%, 18.37\%, and 13.62\% in reading comprehension, domain knowledge, and reasoning, respectively.
Interestingly, we observe that the FG values for bias (Table \ref{bias}) are mostly above zero in the experiments, which suggests that model biases, such as those related to race, color, gender, and so on, are mitigated during continual instruction tuning. For instance, in sentences describing physical appearance, BLOOMZ-7.1b initially prefers stereotype-conforming sentences with a probability of 75.0\%, but this preference decreases to 63.88\% after continual instruction tuning.

\subsection{Effect of Scales}
\label{scalesec}

We visualize the FG values of domain knowledge, reasoning, and reading comprehension with respect to model scales in Figure \ref{DK}. We can observe that the forgetting phenomenon becomes increasingly severe as the model scale increases. For example, the FG values in domain knowledge are 9.54\%, 10.72\%, 14.63\%, and 18.37\% in BLOOMZ-1.1b, 1.7b, 3b, and 7.1b, respectively. BLOOMZ-7.1b suffers the most drastic forgetting. As shown in Table 1, the initial performance $R_o^s$ is boosted by the increasing model scale, but the final performance is relatively similar across different scales, which may explain the varying extent of forgetting. For example the task in domain knowledge are mostly multi-choice problems with 4 options and all the model tend to achieve a nearly random guess performance, but BLLOMZ-7b perform much significant initially. It may imply that these models still shift parameters in a large extent to fit the instruction tasks.
The same pattern can also be observed in the mT0-1.2b and mT0-3.7b models, as shown in Table \ref{main}.
Regarding the bias in LLMs, FG values do not correlate with the model scales, which is also reflected in the initial model performance. In other words, there is no evident correlation between the initial performance $R_o^s$ in bias and the model scales. This finding suggests that the degree of bias in LLMs is not directly related to their size, and increasing the model scale does not necessarily lead to a corresponding increase or decrease in bias.

\subsection{Effect of Model Architecture}
\label{arcsec}

We also compare the forgetting phenomenon of different model architectures in Figure \ref{arc}. As observed, at comparable model scale, BLOOMZ-1.1b and mT0-1.2b achieve similar FG values in domain knowledge, reasoning, and reading comprehension. However, as the scale increases to 3b, BLOOMZ-3b suffers less forgetting compared to mT0-3.7B. For example, the FG value of BLOOMZ-3b is 11.09 which is 5.64 lower than that of mT0-3.7b. These results suggest that BLOOMZ, which has a decoder-only model architecture, can maintain more knowledge during continual instruction tuning. This difference may be attributed to the autoregressive nature of the model or the differences in training objectives. Furthermore, the results imply that as the model scale increases, decoder-only models may suffer from less catastrophic forgetting compared to encoder-decoder models. 
% The detailed results are shown in Figure \ref{arcdetailed}. 
As we observe, the knowledge degraded more drastically in mT0.

\subsection{Effect of General Instruction Tuning}
\label{inssec}

% We also carry out some experiments to analyze the effect of general instruction tuning in the CF phenomenon during continual instruction tuning (Table \ref{tuned}). We compare  BLOOM-7.1b with BLOOMZ-7.1b and LLAMA-7b with ALPACA-7b.  We observe that BLOOMZ-7.1b outperforms BLOOM-7.1b by a large margin in the initial performance of domain knowledge, reasoning, and reading comprehension. Because of the difference in initial performance, BLOOMZ-7.1b suffers a more significant forgetting. However, in LLAMA and ALPACA, there is not a significant gap in the initial performance, and ALPACA maintains more general knowledge after continual fine-tuning. The illustration of the general knowledge is shown in Figure \ref{llama_details}. As we observe that, LLAMA-7b suffers significant forgetting in the first instruction tuning, which implies that the model without general instruction tuning may have less 
% The better-remaining knowledge implies that general instruction tuning can mitigate catastrophic forgetting in LLM during further continual fine-tuning.

We also conduct experiments to analyze the effect of general instruction tuning on the CF problem during continual instruction tuning (Table \ref{tuned}). We compare BLOOM-7.1b with BLOOMZ-7.1b and LLAMA-7b with ALPACA-7B. We observe that BLOOMZ-7.1b outperforms BLOOM-7.1b by a large margin in the initial performance on domain knowledge, reasoning, and reading comprehension. Due to the difference in initial performance, BLOOMZ-7.1b experiences more significant forgetting.
However, in the case of LLAMA and ALPACA, there is no substantial gap in the initial performance, and ALPACA maintains more general knowledge after continual fine-tuning. The illustration of the general knowledge is shown in Figure \ref{llama_details}. We observe that LLAMA-7b suffers significant forgetting in the first instruction tuning, which suggests that models without general instruction tuning may have less ability to retain knowledge during continual fine-tuning. The better retention of knowledge implies that general instruction tuning can mitigate catastrophic forgetting in LLMs during further continual fine-tuning. This finding highlights the importance of general instruction tuning in preserving the acquired knowledge and skills of LLMs when they undergo subsequent task-specific fine-tuning.

\begin{figure*}
    \centering
    \includegraphics[width=\hsize]{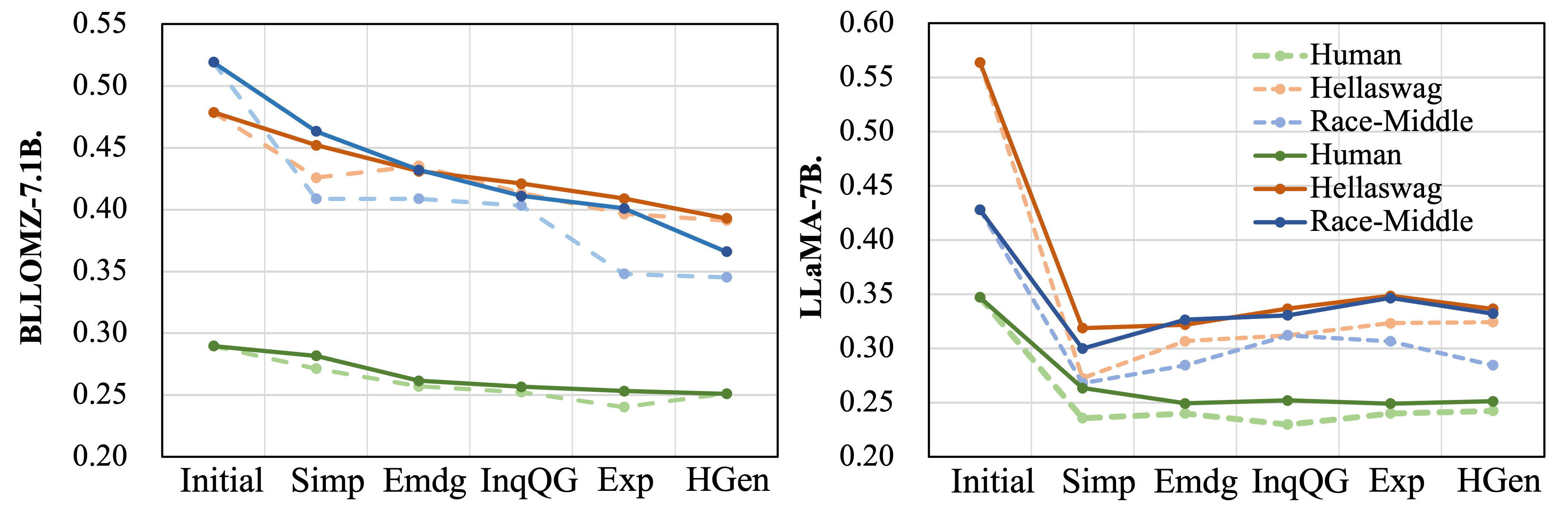}
    % \vspace{-4mm}
    \caption{The performance of general knowledge of the BLOOMZ-7.1b and LLAMA-7b model trained on the instruction data and the mixed data. The dashed lines refers to the performance of BLOOMZ-7.1b and LLAMA-7B and the solid ones refer to those of mixed-instruction trained models.}
    \label{mixed}
        % \vspace{-3mm}
    
\end{figure*}

To further demonstrate the effect of general instruction tuning, 
we mix 10,000 general instruction data samples from ALPACA \citep{alpaca} with the continual instruction tasks to train the BLOOMZ-7b and LLAMA-7b model.
For the sake of brevity, we present the performance of one data split from each evaluation set (MMLU-human, Hellaswag, and Race-middle) to illustrate the effect in Figure \ref{mixed}. 
The results clearly show that the forgetting during continual instruction tuning can be mitigated to a certain extent by incorporating general instruction data. For instance, the performance of MMLU-human in the initial LLAMA-7b model is 34.72\%, but it decreases to 26.8\% when trained solely on the instruction data. However, when trained on the mixed data, the performance becomes 30\%. These findings further further show that general instruction tuning can help alleviate the CF problem encountered during continual instruction tuning.

\section{Conclusion}
In this study, we conducted an empirical investigation into the catastrophic forgetting (CF) phenomenon experienced by large language models (LLMs) during continual instruction tuning. Our findings revealed that the CF problem is generally prevalent in the continual fine-tuning of various LLMs.
Moreover, as the model scale increases, LLMs exhibit a more severe degree of forgetting in domain knowledge, reasoning abilities, and reading comprehension skills. Furthermore, our comparative analysis showed that the decoder-only model, BLOOMZ, demonstrates a superior ability to retain knowledge and skills during continual fine-tuning when compared to the encoder-decoder model, mT0.
Additionally, we discovered that employing general instruction tuning techniques may help alleviate the CF problem in LLMs. Our empirical study suggests that exploring more effective methods to mitigate CF in LLMs during continual fine-tuning is a promising research direction. Meanwhile, since our work is an empirical study and is constrained by the computation resources, there is still large room to investigate the forgetting phenomenon such as a larger model scale (70b or larger) since a larger model may need less parameter changes to fit the downstream tasks.
When applying LLMs, practitioners should remain vigilant and pay close attention to the issue of knowledge forgetting that may occur after instruction tuning. Addressing this challenge is crucial to ensure the reliable and consistent performance of LLMs in real-world applications.

\section{Limitations}
In this study, we take a initial step to analyze the CF problem during continual instruction tuning. Due to the restricted computation resources, we could not carry out experiments on the models with larger scales. But we can still observe the phenomenon  of forgetting from the model scales from 1b to 7b. We control the experiments in a task order for simplifying the analysis, which may affect the forgetting phenomenon. Meanwhile, there are plenty of benchmarks to evaluate the performance of LLMs, here we only adopt some popular ones to analyse the general knowledge, otherwise, the computational cost of conducting experiments would be prohibitively high.

\bibliography{colm2024_conference}
\bibliographystyle{colm2024_conference}

\clearpage
\appendix
\section{Instruction Task Details}
\label{instruction_detail}
We show the instruction samples for the continual instruction tasks adopted in the study in Table \ref{instruction}.

\begin{table*}[h]\small
    \centering
    \begin{tabular}{p{0.9\textwidth}}
        \hline
\textbf{\textit{Simp}}: \\
\textbf{Instruction}: Reformulate this text with simpler words: `His father Robert Alda -LRB- born Alphonso Giuseppe Giovanni Roberto D 'Abruzzo -RRB- was an actor and singer, and his mother Joan Browne was a homemaker and former beauty-pageant winner. ' \\
% \hline
\textbf{Output}: His father, Robert Alda, was an actor, and his mother, Joan, was once crowned Miss New York.  \\
\hline
\textbf{\textit{Emdg}}: \\
\textbf{Instruction}: The associated emotion is `guilty' and the input prompt is `I felt guilty when I was driving home one night and a person tried to fly into my lane, and didn't see me. I honked and they swerved back into their lane, slammed on their brakes, and hit the water cones.'. Now what would be your response, given the following dialogue context:===- Yeah about 10 years ago I had a horrifying experience. It was 100\% their fault but they hit the water barrels and survived. They had no injuries but they almost ran me off the road.  \\
% \hline
\textbf{Output}: No I wasn't hit. It turned out they were drunk. I felt guilty but realized it was his fault. \\
\hline
\textbf{\textit{InqQG}}: \\
\textbf{Instruction}: The del operator shows up in the Laplacian, which is in the kinetic energy term, when the Schrodinger equation is written in the position basis. 2207\^(2) is a differential operator, equal to (d/dx)\^(2) + (d/dy)\^(2) + (d/dz)\^(2), in Cartesian coordinates (the derivatives are partial derivatives).h\^(2)2207\^(2)/2m is the kinetic energy operator in the position basis.=== Given the above text, write the possible curious question it answers:  \\
% \hline
\textbf{Output}: What is the del operator in the Schrodinger equation? \\
\hline

\textbf{\textit{Exp}}: \\
\textbf{Instruction}: Explain why the two following sentences are unrelated: Sentence 1: `This church choir sings to the masses as they sing joyous songs from the book at a church'; Sentence 2: `The church has cracks in the ceiling.'  \\
% \hline
\textbf{Output}: There is no indication that there are cracks in the ceiling of the church. \\
\hline
\textbf{\textit{HGen}}: \\
\textbf{Instruction}: Make a title for this article with `computer': japan's NEC corp. and UNK computer corp. of the United States said Wednesday they had agreed to join forces in supercomputer sales.  \\
% \hline
\textbf{Output}: NEC UNK in computer sales tie-up.  \\
\hline
    \end{tabular}
    \caption{Instruction samples for the continual instruction tasks.}
    \label{instruction}
    \vspace{-2mm}
\end{table*}

\end{document}